\pdfoutput=1

\documentclass[11pt]{article}

\usepackage[]{EACL2023}

\usepackage{inconsolata}

\usepackage{times}
\usepackage{latexsym}

\usepackage[T1]{fontenc}

\usepackage[utf8]{inputenc}

\usepackage{microtype}

\usepackage{hyperref}
\usepackage{url}
\usepackage{graphicx}
\usepackage{booktabs}
\usepackage{multirow, multicol}
\usepackage{diagbox}

\usepackage{ctable}
\usepackage{tabularx}
\usepackage{xspace}
\usepackage{amsmath}
\usepackage{float}
\usepackage{xcolor}
\usepackage{tablefootnote}

\usepackage{pifont}
\usepackage{amssymb}

\usepackage{enumitem}

\definecolor{comment-red}{rgb}{0.8,0,0}

\newcommand{\D}[0]{\mathcal{D}\xspace}

\newcommand{\repr}[0]{Distilled\xspace}

\newcommand{\alignment}[0]{Alignment\xspace}
\newcommand{\trans}[0]{Translation\xspace}
\newcommand{\canon}[0]{Canonicalization\xspace}

\newcommand{\chinese}[0]{Mixed\xspace}

\title{Zero and Few-Shot Localization of Task-Oriented Dialogue Agents \\ with a Distilled Representation}

\author{Mehrad Moradshahi \quad Sina J. Semnani \quad  Monica S. Lam \\
        Computer Science Department \\ Stanford University \\ Stanford, CA \\
        \texttt{\{mehrad,sinaj,lam\}@cs.stanford.edu}}

\begin{document}
\maketitle

\begin{abstract}
\label{sec:abstract}

Task-oriented Dialogue (ToD) agents are mostly limited to a few widely-spoken languages, mainly due to the high cost of acquiring training data for each language. Existing low-cost approaches that rely on cross-lingual embeddings or naive machine translation sacrifice a lot of accuracy for data efficiency, and largely fail in creating a usable dialogue agent.
We propose automatic methods that use ToD training data in a source language to build a high-quality functioning dialogue agent in another target language that has no training data (i.e. zero-shot) or a small training set (i.e. few-shot). Unlike most prior work in cross-lingual ToD that only focuses on Dialogue State Tracking (DST), we build an end-to-end agent.

We show that our approach closes the accuracy gap between few-shot and existing full-shot methods for ToD agents.
We achieve this by (1) improving the dialogue data representation, (2) improving entity-aware machine translation, and (3) automatic filtering of noisy translations.

We evaluate our approach on the recent bilingual dialogue dataset BiToD.
In Chinese to English transfer, in the zero-shot setting, our method achieves 46.7\% and 22.0\% in Task Success Rate (TSR) and Dialogue Success Rate (DSR) respectively. In the few-shot setting where 10\% of the data in the target language is used, we improve the state-of-the-art by 15.2\% and 14.0\%, coming within 5\% of full-shot training.\footnote{Code can be accessed at \url{https://github.com/stanford-oval/dialogues}}




\end{abstract}

\section{Introduction}
\label{sec:intro}

While dialogue agents in various forms have become commonplace in parts of the world, their lack of support for most human languages has prevented access to the benefits they provide for much of the world.
Commercial virtual assistants for example, only support a handful of languages, as extending their functionality to each new language is extremely costly, partially due to the need for collecting new annotated training data in that language.

In recent years, several non-English task-oriented dialogue (ToD) datasets have been created; they are either collected from scratch such as RiSAWOZ~\cite{risawoz} and CrossWOZ ~\cite{zhu-etal-2020-crosswoz}, paraphrased from synthetic sentences by crowdworkers such as BiToD~\cite{lin2021bitod}, or manually translated from another language ~\cite{li2021multi-domain}. All of these approaches are labor-intensive, expensive, and time-consuming; such investment is unlikely to be made for less widely spoken languages. 

Cross-lingual transfer, i.e. using training data from other languages to build a dialogue agent for a specific language, seems especially appealing. An emerging line of work has employed machine translation of training data, and multilingual pre-trained neural networks to tackle this task \cite{ sherborne-etal-2020-bootstrapping, mtop, moradshahi2021contextual}.
However, work in ToD cross-lingual transfer has for the most part, focused on understanding the user input, namely Dialogue State Tracking (DST) and Natural Language Understanding (NLU). Other necessary parts of a dialogue agent like policy and response generation have mostly remained unexplored. 

In this paper, we present a methodology for building a fully functional dialogue agent for a new language (e.g. English), by using  training data in another language (e.g. Chinese) with little to no additional manual dataset creation effort.  
We found that despite prior efforts to improve modeling for existing ToD datasets, the dialogue representation used as input to these models, e.g. full dialogue history in natural language \citep{hosseini2020simple}, is sub-optimal, especially when the training data is either scarce or created automatically using noisy machine translation. We propose a new {\em \repr} representation to fix the shortcomings of current representations.  We also found that previously proposed entity-aware translation technique \citet{moradshahi2021contextual} to be inadequate. Our proposed technique effectively combines entity-aware neural machine translation with text similarity classifiers to automatically create training data for a new language.
This paper explains all the ingredients we found useful, and motivates their use by performing extensive ablation studies.




The contributions of this paper are:

\begin{enumerate}[leftmargin=*,topsep=0pt,noitemsep]
    \item  \emph{A new state-of-the-art result for the BiToD dataset in both few-shot and full-shot settings on English} according to all of our 6 automatic metrics, including an improvement of 14.0\% and 2.9\%,
    respectively, in Dialogue Success Rate (DSR). In fact, using our \repr representation, our few-shot model trained on only 10\% of the training data, achieves similar results to the previous SOTA model trained on 100\% training data.
    
    \item \emph{The first dialogue agent created in the zero-shot cross-lingual transfer setting}, i.e. starting from no training data in the target language. Our agent achieves 71\%, 62\%, 40\%, and 47\% of the performance of a full-shot agent in terms of Joint Goal Accuracy (JGA), Task Success Rate (TSR), DSR, and BLEU score, respectively.
    
    \item 
    \emph{A concise dialogue representation designed for cross-lingual ToD agents}. The \repr dialogue representation works well with our new decomposition of agent subtasks, making significant improvements possible.
    
    \item
    \emph{An improved methodology for automatic translation of ToD training data}.
    We adapt and improve an existing entity-aware machine translation system that localizes entities \cite{moradshahi2021contextual}, extend it to agent response generation, and equip it with a filtering step that increases the quality of the resulting translations.
    


    
\end{enumerate}



\section{Related Work}
\label{sec:related_work}




\subsection{Multilingual Dialogue Datasets}
\label{sec:multilingual-datasets}


MultiWOZ~\cite{multiwoz1, multiwoz2, multiwoz21} and CrossWOZ~\cite{zhu-etal-2020-crosswoz} are two monolingual Wizard-Of-Oz dialogue datasets that cover several domains, suitable for building travel dialogue agents in English and Chinese respectively.
For the 9th Dialog System Technology Challenge (DSTC-9)~\cite{dstc9}, they were translated to Chinese and English using Google Translate. 

GlobalWOZ~\cite{globalwoz}, AllWOZ~\cite{zuo2021allwoz}, and Multi2WOZ~\cite{hung2022multi2woz} translate MultiWOZ to even more languages such as Spanish, Hindi, and Indonesian, with human translators post-editing machine translated dialogue templates, and filling them with newly collected local entities. Although manual post-editing improves data quality and ensures fluency, it also increases the cost and time to create new datasets, thus limiting scalability.


Different from these translation approaches, \citet{lin2021bitod} introduced BiToD, the first bilingual dataset for \emph{end-to-end} ToD modeling. BiToD uses a dialogue simulator to generate dialogues in 5 tourism domains in English and Chinese, then uses crowdsourcing to paraphrase entire dialogues to be more natural. Unlike WOZ-style datasets which usually suffer from poor annotation quality due to human errors~\cite{moradshahi2021contextual}, BiToD is automatically annotated during synthesis.
Since neither manual nor machine translation is used in the creation of BiToD, it does not contain translationese~\cite{eetemadi-toutanova-2014-translationese} or other artifacts of translated text~\cite{clark-tydiqa-2020}, and provides a realistic testbed for cross-lingual transfer of task-oriented dialogue agents. 


\subsection{Multilingual Dialogue State Tracking}
\citet{mrksic-etal-2017-semantic} proposed using cross-lingual word embeddings for zero-shot cross-lingual transfer of DST models.
With the advent of large language models, contextual embeddings obtained from pre-trained multilingual language models~\cite{devlin2018bert, mt5, liu2020multilingual} have been used to enable cross-lingual transfer in many natural language tasks, including DST. 

\citet{chen-etal-2018-xlnbt} used knowledge distillation~\cite{hinton2015distilling} to transfer DST capabilities from a teacher DST model in the source language to a student model in the target language.

Machine translation has been used for DST, both as a way of obtaining cross-lingual representations, and to translate training data. For instance, \citet{schuster-etal-2019-cross-lingual} used representations obtained from machine translation models and reported that it performs better than training with machine translated training data for single-turn commands.
More advanced data translation approaches like the entity-aware method of \citet{moradshahi2021contextual} further improved the DST data quality achievable with machine translation.

\section{Distilled ToD Agent}
Our methodology includes a dialogue task decomposition and a \repr dialogue representation that are tailored to cross-lingual ToD agents.
In this section we describe these two components.

We follow the end-to-end task-oriented dialogue (ToD) setting~\cite{hosseini2020simple} where a user converses freely with an agent over several turns to accomplish his/her goal with all of its constraints (e.g. ``book a restaurant that is rated at least 3.''). In each turn, the agent must access its database if needed to find the requested information (e.g. find a restaurant that satisfies user constraints), decide on an action (e.g. to present the information to the user or to ask follow-up questions) and finally respond to the user in natural language based on the action it selects.

\begin{figure}
\centering
\includegraphics[width=\linewidth]{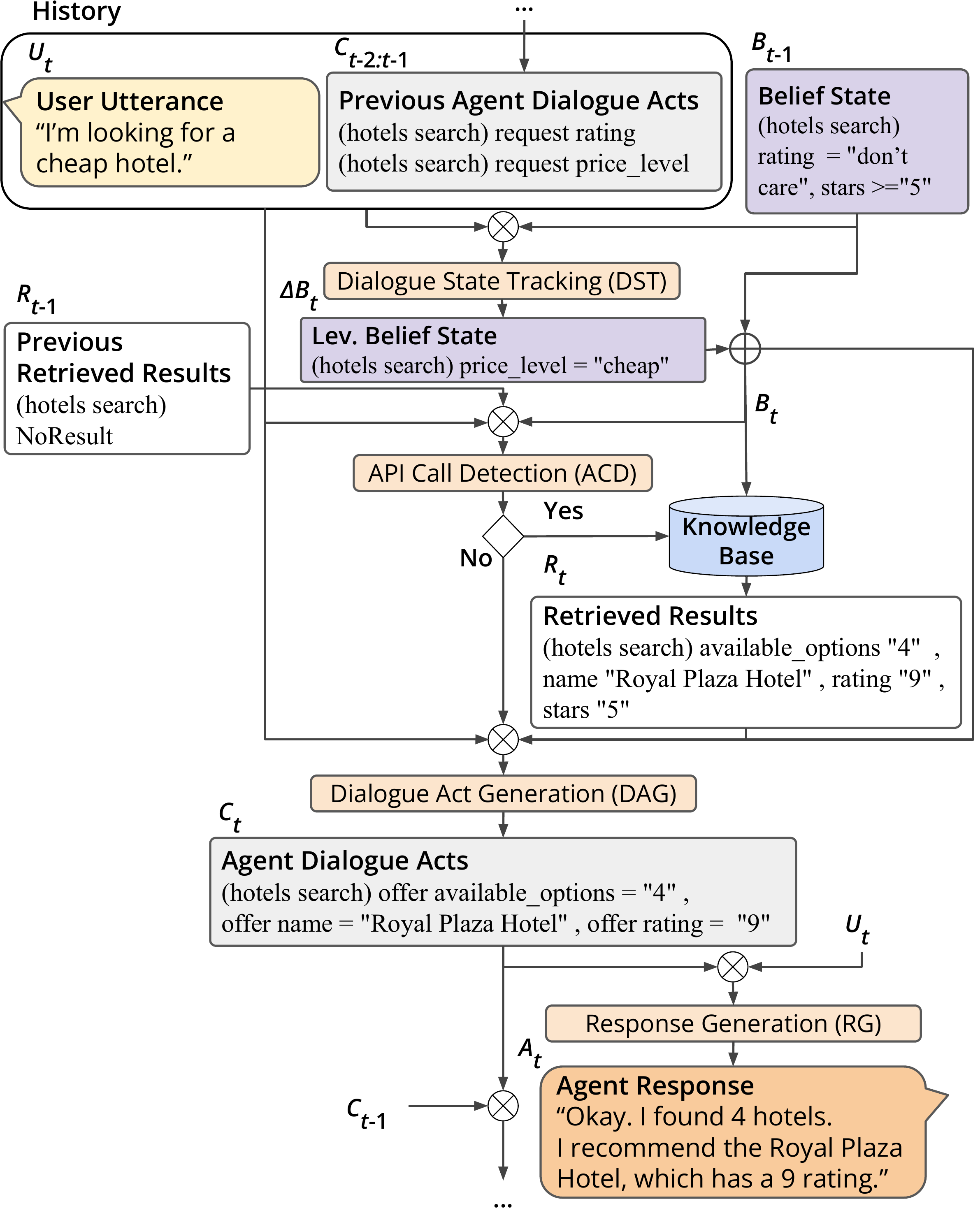}
\vspace{-1em}
\caption{Inference-time flow diagram for our dialogue agent. DST, ACD, DAG, and RG share the same neural model. $U$, $A$, $C$, $B$, and $R$ indicate user utterance, agent response, agent dialogue acts, dialogue state, and retrieved database results respectively. $t$ is the turn number. $\otimes$ indicates text concatenation. $\oplus$ refers to the update rule in Equation~\ref{eq:dst}. 
}
\label{fig:dialogue-loop}
\vspace{-1.6em}
\end{figure}


\subsection{Preliminaries}
Formally, a \emph{dialogue} $D=\{U_1, A_1, ..., U_T, A_T\}$ is a set of alternating user utterances $U_t$ and agent responses $A_t$ for a number of turns $T$.


A \emph{belief state} at turn $t$, $B_t$, consists of a list of $\langle \text{\em domain}, \text{\em intent} \rangle$ tuples and a set of $\langle \text{\em slot}, \text{\em relation}, \text{\em value} \rangle$ tuples. \emph{Intent} is the user intent, either search or book.
\emph{Relation} is a comparison or membership operator. \emph{Value} can be one or more entity names or strings from the ontology, or a literal.  To see all possible domains, slots and values please refer to Table 4 in~\citet{lin2021bitod}.


The \emph{Levenshtein belief state}~\cite{lin-etal-2020-mintl} is the difference between belief states in consecutive turns, i.e. $\Delta B_t = B_t-B_{t-1}$. It captures only the relations and values that have changed in the last user utterance, or tuples that have been added or removed.




An \emph{Agent dialogue act} at turn $t$, $C_t$, is a list of $\langle \text{\em domain},\text{\em intent}\rangle$ tuples and a set of $\langle \text{\em dialogue\_act\_name}, \text{\em slot}, \text{\em value}\rangle$ tuples indicating the action the agent takes and the information offered to the user, if any.


\subsection{Task Decomposition}
The task of dialogue agents is usually broken down to subtasks, which may be performed by a pipelined system~\cite{gao2018neural} or by a single neural network~\cite{hosseini2020simple, lei2018sequicity}. Here we describe our subtasks and their inputs and outputs (Figure \ref{fig:dialogue-loop}).


After the user speaks at turn $t$, the agent has access to the belief state up to the previous turn ($B_{t-1}$), the history of agent dialogue acts ($C_1, ..., C_{t-1}$), and the history of agent and user utterances so far ($A_1, ..., A_{t-1}$ and $U_1, ..., U_{t}$). 
Our agent performs the following four subtasks:

\begin{enumerate}[leftmargin=*,topsep=0pt,noitemsep]
    \item
    \emph{Dialogue State Tracking (DST)}: Generate $\Delta B_t$, the Levenshtein  belief state, for the current turn based on the previous belief state, the last two agent dialogue acts\footnote{Our ablation study described in Section \ref{sec:ablation} justifies the use of the last two agent dialogue acts instead of just the last one.}, and the current user utterance. $\Delta B_t$ is combined with $B_{t-1}$ to produce the current belief state. 
    
    \vspace{-1.5em}
    \begin{equation}
        \label{eq:dst}
        \begin{aligned}
        &\Delta B_t = \mathrm{DST}(B_{t-1}, C_{t-2}, C_{t-1}, U_{t})\\
        &B_{t} \leftarrow B_{t-1} + \Delta B_t
        \end{aligned}
    \end{equation}
    \vspace{-1em}
    
    \item
    \emph{API Call Detection (ACD)}: Call an API to query the database, if needed.  

    \vspace{-1.5em}
    \begin{equation}
        \label{eq:api-call?}
        q_t = \mathrm{ACD}(B_t, C_{t-2}, C_{t-1}, U_{t}, R_{t-1})
    \end{equation}
    \vspace{-2em}
      \begin{equation}
        \label{eq:api}
        R_t \; \leftarrow \; q_t ? \; \mathrm{KB}(B_t) \; : \; \emptyset
    \end{equation}
    \vspace{-1em}
      
    In turn $t$, ACD determines if an API call is necessary. If so, the result $R_t$ is the top entity in the knowledge base KB, based on a deterministic ranking scheme, that matches the API call constraints in $B_t$, and is empty otherwise. If no entities match the constraint, we set $R_t$ to the special value {\sc NoResult}.

    
    \item
    \emph{Dialogue Act Generation (DAG)}: Generate  $C_t$, the agent dialogue act for the current turn based on the current belief state, the last two agent dialogue acts, the user utterance, and the result from the API call. 
    
    \vspace{-2em}
    \begin{equation}
        \label{eq:da}
        C_t = \mathrm{DAG}(B_t, C_{t-2}, C_{t-1}, U_{t}, R_t)
    \end{equation}
    \vspace{-1.3em}
    
    \item
    \emph{Response Generation (RG)}: Convert the agent dialogue act $C_t$ to the new agent utterance $A_t$. Note that $C_t$ contains all the necessary information for this subtask. However, providing $U_t$ improves response fluency and choice of words, leading to a higher BLEU score, partly due to mirroring~\cite{t2g2}.
    
    \vspace{-1em}
    \begin{equation}
        \label{eq:eq4}
        A_t = \mathrm{RG}(U_t, C_t)
    \end{equation}
    \vspace{-2em}
    
\end{enumerate}

\subsection{The \repr Dialogue Representation}
\label{sec:representation}

The design of \repr is based on the following principles:

\begin{enumerate}[leftmargin=*,topsep=0pt,noitemsep]
    \item 
    For cross-lingual agents, it is important to reduce the impact of translation errors.  
    The representation should make minimal use of natural language by using a formal representation where possible. 
    
    \item
    Dialogues can get long, but the representation should be succinct, containing only the necessary information, so the neural network need not \emph{learn} to ignore unnecessary information from copious data. This improves data efficiency as well as the training and inference speed of neural models.
   
\end{enumerate}

We note that BiToD’s original representation~\cite{lin2021bitod} follows neither of these principles.\footnote{We found this to be true for several previously-proposed popular representations of MultiWOZ as well~\cite{lei2018sequicity, chen2019semantically}.}
It makes extended use of natural language: all previous user and agent natural language utterances are included in the input of all subtasks.
It has many redundancies: for each subtask, it inputs the concatenation of all previous subtask's inputs and outputs. In the following, we highlight the changes we made to the~\cite{lin2021bitod} representation. 




\paragraph{Replace agent utterances with formal agent dialogue acts.}
Since agent responses are automatically generated, it is possible to capture all information useful to the different subtasks with formal agent dialogue acts. In this way, the neural network need not interpret previous natural language utterances.

We take two steps to generate the agent responses: DAG (Dialogue Act Generation) first produces the formal act, $C_t$, which is then fed into RG (Response Generation) to generate the natural language response $A_t$. Note that RG is not a part of the dialogue loop: the natural language $A_t$ only serves to communicate to the user; it is the formal $C_t$ from DAG that gets fed to subsequent subtasks instead. In contrast, \citet{lin2021bitod} generates the agent response directly from API results. ~\citet{hosseini2020simple} also separates the response generation into two steps, but they use $A_t$ instead of $C_t$ as input to the semantic parser for the next turn. 

Note that the agent dialogue acts are independent of the natural language used in the dialogues, if we ignore the entity values.  This is beneficial to cross-lingual agents as it can learn easier from data available in other languages. 
Furthermore, DAG can be validated on whether the output dialogue acts match the gold answers exactly.  This is not possible with natural language results, whose quality is typically estimated with BLEU score. 
    
\paragraph{Shorten user utterance history.}
Since the belief state formally summarizes what the user has said, we remove previous user utterances $U_1, ..., U_{t-1}$ from input to all subtasks, relying on the belief state $B_{t-1}$ instead. 


 
\paragraph{Untangle API call detection from response generation.}
    After DST is done, depending on whether or not an API call is needed, \citet{lin2021bitod} either directly generates the agent response, or makes the API call and then generates the response in two steps. Our design is to always take two steps: (1) generate the API call \emph{or indicate that there is none}, and (2) generate the agent response. 
\section{Automatic Dialogue Data Translation}
\label{sec:data}

Given a training dataset for one language, we automatically generate a training set in the target language we are interested in. This problem has been studied in the context of NLU for questions~\cite{moradshahi-etal-2020-localizing,sherborne-etal-2020-bootstrapping,mtop} and for dialogues~\cite{moradshahi2021contextual,globalwoz,zuo2021allwoz}. One challenge is that the translated dataset should refer to entities in the target language. Thus, \citet{moradshahi-etal-2020-localizing} proposed to first use cross-attention weights of the neural translation model to align entities in the original and translated sentences, then replace entities in the translated sentences with local entities from a target language knowledge base. Our initial experiments showed that applying this approach directly to end-to-end dialogue datasets does not yield good performance, especially for response generation. Thus, we adapted and improved this approach for dialogues as discussed below.

\subsection{Alignment for Dialogues}
First, we found that while translation with alignment works for NLU, it does not work well for RG. Machine translation introduces two kinds of error: (1) Translated sentences can be ungrammatical, incorrect, or introduce spurious information. (2) The alignment for entities may be erroneous, which can seriously hurt the factual correctness of the responses. As shown in ~\citet{moradshahi2021contextual}, these errors are tolerable in NLU since (1) sentences are seen by machines, not shown to users, (2) pre-trained models like mBART are somewhat robust to noisy inputs, since they are pre-trained on perturbed data.
However, training with such low-quality data is not acceptable for RG, since the learned responses are shown directly to the user. 

Second, we found alignment recall to be particularly low for an important category: entities that are mostly quantitative. We observe that dates, times, and prices can be easily mapped between different languages using rules. We propose to first try to translate such entities with dictionaries such as those available in dateparser~\cite{dateparser} and num2words~\cite{num2words}, and to match them in the translated text. We resort to using neural alignment only if no such match is found.

\subsection{Filtering Translation Noise for RG}
\label{subsec:filter}

To reduce translation noise for RG, we automatically filter the translated data based on the semantic textual similarity between the source and translated sentences. For this purpose, we use LaBSE~\cite{feng2020language}, a multilingual neural sentence encoder based on multilingual BERT~\cite{devlin2018bert}, trained on translation pairs in various languages with a loss function that encourages encoding pairs to similar vectors. To score a pair of sentences, the model first calculates an embedding for each sentence and computes the cosine distance between those vectors. The lower the distance is, the more semantically similar the sentences are, according to the model. 

In creating the RG training set, we first translate the source agent utterances to the target language and use LaBSE to remove pairs whose similarity score is below a threshold. We found a threshold of 0.8 to work best empirically. Higher thresholds would inadvertently filter correctly translated utterances. We construct the final training data by pairing aligned translated utterances that pass the filter with their corresponding translated agent dialogue acts.




\section{Experiment Setting}
\label{sec:experiments}

\subsection{Base Dataset}
We perform our experiments on BiToD, a large-scale high-quality bilingual dataset created using the Machine-to-Machine (M2M) approach. 
It is a multi-domain dataset, including restaurants, hotels, attractions, metro, and weather domains. It has a total of 7,232 dialogues (3,689 dialogues in English and 3,543 dialogues in Chinese) with 144,798 utterances in total. The data is split into 5,787 dialogues for training, 542 for validation, and 902 for testing. The training data is from the same distribution as validation and test data.

\subsection{Implementation details}
\label{sec:implementation}
Our code is implemented in PyTorch~\cite{paszke2019pytorch} using GenieNLP~\cite{geniepldi19} library for training and evaluation metrics. We also use the Dialogues\footnote{\url{https://github.com/stanford-oval/dialogues}} library for data preprocessing and evaluation. We use pre-trained models available through HuggingFace's Transformers library~\cite{Wolf2019HuggingFacesTS}. The following model names are from that library.
We use \textit{mbart-large-50} as the neural model for our agent in all our experiments.
All models use a standard Seq2Seq architecture with a bidirectial encoder and left-to-right autoregressive decoder. mBART is pre-trained to denoise text in 50 languages, while mT5 is trained on 101 languages. mBART uses sentence-piece~\cite{kudo2018sentencepiece} for tokenization.

In each setting, all four subtasks of DST, API detection, dialogue act generation, and response generation are done in a single model, where we specify the task by prepending a special token to the input. We found mBART to be especially effective in zero-shot settings as the language of its outputs can be controlled by providing a language-specific token at the beginning of decoding. Additionally, its denoising pre-training objective improves its robustness to the remaining translation noise.

For translation, we use the publicly available \textit{mbart-large-50-many-to-one-mmt} (\textasciitilde611M parameters) model which can directly translate text from any of the 50 supported languages to English. It is an mBART model additionally fine-tuned to do translation. We use greedy decoding and train our models using teacher-forcing and token-level cross-entropy loss. We used Adam~\cite{kingma2014adam} as our optimizer with a starting learning rate of $2 \times 10^{-5}$ and linear scheduling. These hyperparameters were chosen based on a limited hyperparameter search on the validation set. For the numbers reported in the paper, due to cost, we performed only a single run for each experiment.
 
Our models were trained on virtual machines with a single NVIDIA V100 (16GB memory) GPU on the AWS platform. For a fair comparison, all monolingual models were trained for the same number of iterations of 60K, and bilingual models for 120K. In the few-shot setting, we fine-tuned the model for 3K steps on 1\% of the data and 6K steps on 10\% of the data.
Sentences are batched based on their input and approximate output token count for better GPU utilization. We set the total number of tokens per batch to 800 for mBART. Due to the verbosity and redundancy of the original BiToD representation, ~\citet{lin2021bitod} used a batch size of 1 example for training mbart-large. Using our \repr representation, however, we can fit up to ~6 examples in each batch and process each batch 3 times faster during training. Training and evaluating each model takes about 10 GPU-hours on average.
 
During error analysis, we noticed that although certain slots (max\_temp and min\_temp slots in Metro domain, and time and price\_range slots in Weather domain) are present in the retrieved knowledge base values, the model does not learn to output them in the agent dialogue act generation subtask. This issue stems from BiToD's non-deterministic policy where an agent sometimes provides these slots and sometimes not in the gold training data. To mitigate this, during evaluation, we automatically check if these slots are present in the input and append them and their retrieved values to the generated agent dialogue acts. 



At inference time, we use the predicted belief state as input to subsequent turns instead of ground truth. However, to avoid the conversation from diverging from its original direction, ~\citet{lin2021bitod} use the ground-truth natural-language agent response as input for the next turn.  To make sure the settings are equivalent for a fair comparison, we use ground-truth agent acts as input for the next turn.

\subsection{Evaluation Metrics}

We use the following metrics to compare different models. Scores are averaged over all turns unless specified otherwise.

\begin{itemize}[leftmargin=*,topsep=3pt,itemsep=3pt]
    \item
    \textbf{Joint Goal Accuracy (JGA)}~\cite{multiwoz1}: Is the standard metric for evaluating DST. JGA for a dialogue turn is 1 if all slot-relation-value triplets in the generated belief state match the gold annotation, and is 0 otherwise.
    
    \item
    \textbf{Task Success Rate (TSR)}~\cite{lin2021bitod}: A task, defined as a pair of domain and intent, is completed successfully if the agent correctly provides all the user-requested information and satisfies the user's initial goal for that task. TSR is reported as an average over all tasks. 
    
    \item
    \textbf{Dialogue Success Rate (DSR)}~\cite{lin2021bitod}: DSR is 1 for a dialogue if all user requests are completed successfully, and 0 otherwise. DSR is reported as an average over all dialogues. We use this as the main metric to compare models, since the agent needs to complete all dialogue subtasks correctly to obtain a full score on DSR.
    
    \item
    \textbf{API}~\cite{lin2021bitod}: For a dialogue turn, is 1 if the model correctly predicts to make an API call, and all the constraints provided for the call match the gold. It is 0 otherwise. 
    
    \item
    \textbf{BLEU}~\cite{papineni2002bleu}: Measures the natural language response fluency based on n-gram matching with the human-written gold response. BLUE is calculated at the corpus level.
    
    \item
    \textbf{Slot Error Rate (SER)}~\cite{wen2015semantically}: It complements BLEU as it measures the factual correctness of natural language responses. For each turn, it is 1 if the response contains all entities present in the gold response, and is 0 otherwise.
\end{itemize}

\begin{table*}[!htb]
\small
\centering
\newcolumntype{C}{>{\centering\arraybackslash}X}
\begin{tabularx}{\linewidth}{l|C|C|C|C|C|C}
\specialrule{.1em}{.0em}{.0em} 

Representation & JGA $\uparrow$ & TSR $\uparrow$  & DSR $\uparrow$  & API $\uparrow$  & BLEU $\uparrow$  & SER $\downarrow$ \\
\specialrule{.1em}{.0em}{.0em} 
Original~\cite{lin2021bitod}   & 69.19 & 69.13 & 47.51 & 67.92 & 38.48 & 14.93  \\

\repr (Ours)   & \bf 76.79 & \bf 75.64 & \bf 53.39 & \bf 76.33 & \bf 42.54 & \bf 10.61 \\
{\tiny $\bullet$} Generate full state     & 74.30 & 74.19 & 50.90 & 73.93 & 41.90 & 11.38 \\
{\tiny $\bullet$} Natural agent response     & 75.62 & 73.41 & 49.10 & 73.93 & 40.94 & 11.90  \\
{\tiny $\bullet$} Only last agent turn   & 73.97 & 74.19 & 52.71 & 74.27 & 41.83 & 11.81 \\
{\tiny $\bullet$} Prev. user utterance as state    & 71.75 & 61.66 & 33.94 & 67.67 & 39.72 & 15.97 \\
{\tiny $\bullet$} Remove state  & 70.84 & 51.89 & 24.43 & 66.47 & 37.10 & 19.61 \\

\specialrule{.1em}{.0em}{.0em} 
\end{tabularx}
\vspace{-0.5em}
\caption{Full-shot English monolingual training with ablation. All results are reported on the English test set of BiToD using the same evaluation script. The best result is in bold.} 
\label{tab:ablations}
\vspace{-1.5em}
\end{table*}

\section{Results and Discussion}


We first show how our \repr representation affects the performance of an agent in a full-shot setting. We then evaluate our proposed techniques on cross-lingual settings with varying amounts of available training data.

\subsection{Evaluation of \repr Representation}
\label{sec:ablation}
To understand how our design of \repr representation affects the performance of ToD agents in general, we train an English agent using all the English training data and perform an ablation study (Table \ref{tab:ablations}). We observe that even though the \repr representation removes a lot of natural language inputs, it improves the best previous English-only results on JGA, TSR, DSR, API, BLEU and SER by 7.6\%, 6.5\%, 5.9\%, 8.4\%, 4.1\%, and 4.7\%, respectively.
This suggests that natural language utterances carry a lot of redundant information, and the verbosity may even hurt the performance.
Note that the improvement in BLEU is also accompanied by an improvement of factuality measured by SER.

Furthermore, using the \repr representation reduces training time by a factor of 3. See Section~\ref{sec:implementation} for more details.  

\paragraph{Generate full state.} 
Our first ablation study confirms that the proposal by ~\citet{lin-etal-2020-mintl} to predict the Levenshtein belief state ($\Delta B_t$) is indeed better than the cumulative state ($B_t$). Note that the training time per gradient step is more than twice as long in this ablation since the outputs are longer. 


\paragraph{Natural agent response.} Here we use natural language agent responses as input instead of agent dialogue acts, replacing $C_{t-1}, C_{t-2}$ with $A_{t-1}, A_{t-2}$. The drop in TSR and DSR shows this is an important design choice - distilling natural language into a concise formal representation improves the model's ability to understand the important information in the sentence.  




\paragraph{Only last agent turn.} When we remove $C_{t-2}$ from the input and only use $C_{t-1}$, we observe a drop across all metrics. This is because some turns in BiToD refer to the agent's states from two turns ago. We experimented with carrying three turns, but there was no improvement. 


\paragraph{Previous user utterance as state.}
In this ablation, we use $U_{t-1}$ instead of $B_{t-1}$ as subtask inputs. Compared to all previous ablations, accuracy drastically decreases across all metrics, especially JGA. This is expected since the information from earlier turns present in the dialogue state is now lost. Additionally, it shows that the dataset is highly contextual and therefore a summary of the conversation history is necessary.

\paragraph{Remove state.} We remove $B_{t-1}$ without adding back the previous user utterance $U_{t-1}$. Compared to the previous ablation, TSR and DSR drop by 10.5\% and 5.2\% respectively. This difference shows $U_{t-1}$ does contain part of the information captured in $B_{t-1}$.


\subsection{Evaluation of Cross-Lingual Transfer}
The goal of this experiment is to create an agent in a {\em target} language, given the full training data in a source language ($\D_\mathrm{src}$), and a varying amount of training data in a target language ($\D_\mathrm{tgt}$). We also assume that valuation and test data are available in both source and target languages.  We chose Chinese as the source language and English as the target language so we can perform error analysis and the model outputs are understandable for a wider audience. 

\subsubsection{Varying Target Training Data}
\paragraph{Full-Shot.}
In the full-shot experiments, all of $\D_\mathrm{tgt}$ is available for training.  We train two models on two data sets: (1) on a shuffled mix of $\D_\mathrm{src}$ and $D_\mathrm{tgt}$. 
(2) on $\D_\mathrm{tgt}$ alone. The ablation ``\textit{$-$\chinese}'' in Table~\ref{tab:accuracies} refers to the latter.

\begin{table*}[!htb]
\small
\centering
\newcolumntype{C}{>{\centering\arraybackslash}X}
\begin{tabularx}{\linewidth}{l|C|C|C|C|C|C}

\specialrule{.1em}{.0em}{.0em} 

Setting & JGA $\uparrow$ & TSR $\uparrow$  & DSR $\uparrow$  & API $\uparrow$  & BLEU $\uparrow$  & SER $\downarrow$ \\

\specialrule{.1em}{.0em}{.0em} 
\multicolumn{7}{c}{Full-Shot} \\
\specialrule{.1em}{.0em}{.0em} 
MinTL(mT5) & 72.16 & 71.18 & 51.13 & 71.87 & 40.71 & 13.75  \\
$-$ \chinese & 69.19 & 69.13 & 47.51 & 67.92 & 38.48 & 14.93 \\
\hline
MinTL(mBART)  & 69.37 & 42.45 & 17.87 & 65.35 & 28.76 & -- \\
$-$ \chinese & 67.36 & 56.00 & 33.71 & 57.03 & 35.34 & --   \\
\hline
Ours    & \bf 77.52 & 75.04 & \bf 54.07 & 74.44 & 41.46 & 11.17 \\
$-$ \chinese  & 76.79 & \bf 75.64 & 53.39 & \bf 76.33 & \bf 42.54 & \bf 10.61 \\

\specialrule{.1em}{.0em}{.0em} 
\multicolumn{7}{c}{Zero-Shot} \\
\specialrule{.1em}{.0em}{.0em} 
Ours   & \bf 55.33 & \bf 46.74 & \bf 21.95 & \bf 63.04 & \bf 20.01 & \bf 20.52 \\
$-$ Filtering    & 54.83 & 45.03 & 19.68 & 60.81 & 19.11 & 20.86 \\
~$-$ Alignment  & 47.21 & 4.72 & 1.13 & 52.74 & 8.26 & 39.20 \\
~~$-$ Translation  & 14.73 & 3.52 & 1.58 & 6.26 & 0.69 & 41.30 \\
~~~$-$ Canonicalization    & 2.13 & 1.20 & 0.00 & 0.26 & 0.25 & 42.39 \\

\specialrule{.1em}{.0em}{.0em} 
\multicolumn{7}{c}{Few-Shot (1\%)} \\
\specialrule{.1em}{.0em}{.0em} 
Ours   & \bf 64.60 & \bf 57.89 & \bf 34.16 & \bf 62.09 & \bf 28.15 & \bf 17.94 \\
$-$ Filtering    & 63.88 & 57.80 & 32.35 & 59.95 & 28.00 & 18.57 \\
~$-$ Alignment   & 58.86 & 51.89 & 23.76 & 57.12 & 26.84 & 21.56  \\
~~$-$ Translation   & 49.58 & 41.34 & 19.68 & 46.05 & 22.73 & 24.86  \\
~~~$-$ Canonicalization  & 44.56 & 42.97 & 20.36 & 46.23 & 23.08 & 24.77  \\
Few-shot Only   & 25.08 & 24.61 & 11.09 & 23.67 & 18.71 & 32.62 \\

\specialrule{.1em}{.0em}{.0em} 
\multicolumn{7}{c}{Few-Shot (10\%)} \\
\specialrule{.1em}{.0em}{.0em} 
MinTL(mT5) & 58.85 & 56.43 & 34.16 & 57.54 & 31.20 & -- \\
$-$ Translation & 48.77 & 44.94 & 24.66 & 47.60 & 29.53 & 19.75 \\
Few-shot Only  & 19.86 & 6.78 & 1.36 & 17.75 & 10.35 & -- \\
\hline
MinTL(mBART)   & 37.50 & 21.61 & 10.18 & 27.44 & 17.86 & -- \\
$-$ Translation  & 42.84 & 36.19 & 16.06 & 41.51 & 22.50 & -- \\
Few-shot Only  & 4.64 & 1.11 & 0.23 & 0.60 & 3.17 & -- \\
\hline

Ours    & \bf 72.70 & \bf 71.61 & \bf 48.19 & \bf 72.56 & \bf 36.02 & \bf 12.71 \\
$-$ Filtering    & 72.45 & 69.55 & 44.57 & 69.55 & 34.67 & 13.62 \\
~$-$ Alignment    & 68.40 & 63.38 & 38.24 & 63.38 & 32.99 & 16.63  \\
~~$-$ Translation  & 67.13 & 63.12 & 41.40 & 63.64 & 32.86 & 16.40  \\
~~~$-$ Canonicalization   & 64.51 & 63.64 & 40.27 & 62.69 & 32.71 & 16.63 \\
Few-shot Only    & 57.18 & 54.80 & 28.73 & 55.66 & 29.61 & 19.66 \\

\specialrule{.1em}{.0em}{.0em}
\end{tabularx}
\vspace{-0.5em}
\caption{All results are reported on the original English test set of BiToD using the same evaluation script. The best result in each section is in bold. Each ``$-$'' removes one additional component from the previous row. All MinTL results are from \citet{lin2021bitod}. SER numbers are not available for some models. An upward arrow is show for columns where bigger numbers are better, and vice versa.}
\label{tab:accuracies}
\vspace{-1.5em}
\end{table*}

\paragraph{Zero-Shot.}
In our zero-shot experiments, we train with a canonicalized $\D\mathrm{src}$ and an automatically translated data set, as explained below.

    \textit{Canonicalization}: To increase transfer learning from the source to the target language, we use the same canonical formal representation across languages~\cite{moradshahi-etal-2020-localizing,razumovskaia2021crossing}. 
    To do so, we adapt $\D\mathrm{src}$ so that the domain names, slot names, agent dialogue acts, and API names in the formal representation to be the same as the target language. Note that the user utterance, agent response, and slot values will remain in the source language. The BiToD dataset has a one-to-one mapping for most of the above and we added the missing items. 

    \textit{Translation}: We use machine translation to convert the user and agent utterances and slot values in $\D_\mathrm{src}$ to create a training set for the target language. 
    
    \textit{Alignment}: After translating the data, we use alignment (Section~\ref{sec:data}) to localize entities while ensuring the entities in translated utterances still match the values specified in annotations. 

    \textit{Filtering}: We use the filtering procedure described in Section~\ref{subsec:filter} to remove turns where agent responses are deemed to have low translation quality.

In Table~\ref{tab:accuracies}, \textit{Ours} refers to our main approach, which combines all four techniques. Each ablation incrementally takes away one of the techniques.

\paragraph{Few-Shot.}
In the few-shot setting, we start with our pre-trained zero-shot models (with various ablations) and further fine-tune it on 1\% and 10\% of $\D_\mathrm{tgt}$, which comprises 29 and 284 dialogues, respectively. \citet{lin2021bitod} reported the results only for the 10\% setting. We use their few-shot data split in that case to be directly comparable. We add one more ablation study where we eliminate cross-lingual transfer by training a model only on the few-shot data (Few-shot Only).


\subsubsection{Baseline}
We compare our results to the best previously reported result on BiToD from \citet{lin2021bitod}.
This SOTA result was obtained using MinTL~\cite{lin-etal-2020-mintl} and using a single mT5-small model to perform all dialogue subtasks.

Contrary to what \citet{lin2021bitod} reported, we found that mBART-large model outperforms mT5-small in all settings. We have included all the results including MinTL(mBART) in Table~\ref{tab:accuracies} for comparison.

\subsubsection{Results}
The results for our cross-lingual experiment are reported in Table~\ref{tab:accuracies}. 
Overall, in the full-shot setting, when training on both source and target language data, we improve the SOTA in JGA by 5.3\%, TSR by 3.8\%, DSR by 2.9\%, API by 2.6\%, BLEU by 0.8\%, and SER by 2.6\%.

Our zero-shot agent achieves 71\%, 62\%, 40\%, and 47\% of the performance of a full-shot agent in terms of JGA, TSR, DSR, and BLEU score, respectively.
In the 10\% few-shot setting, our approach establishes a new SOTA by increasing JGA, TSR, DSR, API, and BLEU absolutely by 13.9\%, 15.2\%, 14.0\%, 15.0\%, and 4.8\% respectively. Prominently, training with just 10\% of the data beats the full-shot baseline which is trained on 100\% of the training data, on all metrics except for DSR and BLEU. It also comes within 5\% of full training using the \repr representation on all metrics.


\emph{Our \repr representation improves the performance, especially in few-shot}. Comparing our results with that of \citet{lin2021bitod}, in the full-shot monolingual setting (MinTL(mT5) ``$-$Mixed'' vs. Ours ``$-$Mixed''), models trained on data with our representation outperform the baseline on all metrics.
In the pure few-shot (10\%) setting, Ours outperforms MinTL(mT5) significantly in all metrics. This suggests that our \repr representation and task decomposition are much more effective in low-data settings.




\emph{Canonicalization is useful.} Comparing ``$-$\trans'' with ``$-$\canon'', training on canonicalized data significantly improves the results in the zero-shot setting. This is intuitive since canonicalization brings training data closer in vocabulary to the test data in the target language. This improvement comes at almost no cost since translation is done automatically using a dictionary.

\emph{Automatic naive translation of the training set does not work for zero-shot.} 
 The naive translation approach (i.e. without alignment) completely fails in the zero-shot setting by achieving only 4.7\% in TSR, and 1.1\% in DSR, as translated entities might no longer match with ones in the annotation.  Adding few-shot data helps significantly as the gap closes between ``$-$\alignment'' and ``$-$\trans'' ablations.

\emph{Alignment improves translation quality in all settings and metrics.}
With alignment, the translation approach performs much better in all settings, establishing a new state-of-the-art in zero and few-shot settings according to almost all metrics. As a general trend, the lower data settings benefit more from alignment.
We additionally performed an experiment using the alignment proposed by ~\cite{moradshahi2021contextual}. There is a 4.0\% drop in TSR and 4.5\% in DSR, confirming the benefit of our improved alignment.  

\emph{Filtering noise for RG improves fluency.}
We perform an ablation by training separate models on filtered and unfiltered translated agent utterances. The filtering process is described in Section~\ref{subsec:filter}.
In 10\% fewshot setting, both BLEU and SER improve by 1.4\% confirming that automatically removing poor translations from training data improves the agent response quality. Interestingly, we observe an increase in other metrics too. Since model parameters are shared between all subtasks, enhancing the data quality for one subtask will have a positive impact on the others as well.

\section{Conclusion}
\label{sec:conclusion}

This paper shows how to build a dialogue agent in a new language automatically, given a dialogue dataset in another language, by using entity-aware machine translation and our new \repr dialogue representation. The performance can be further improved if a few training examples in the target language are available, and we show that our approach outperforms existing ones in this setting as well. 

On the BiToD dataset, our method achieves 3.9\% and 2.9\% improvement in TSR and DSR, respectively, over the previous SOTA in full-shot setting, and 15.2\% and 14.0\% in a 10\% few-shot setting, showing the effectiveness of our approach. More importantly, training on translated data and only 10\% of original training data comes within 5\% of full training.

We have implemented our methodology as a toolkit for developing multilingual dialogue agents, which we have released open-source. Our proposed methodology can significantly reduce the cost and time associated with data acquisition for task-oriented dialogue agents in new languages.

\section{Limitations}

As discussed in Section~\ref{sec:multilingual-datasets}, organic (i.e. without the use of translation) multilingual dialogue datasets are scarce, which has limited the scope of our experiments. Our guidelines to improve dialogue representation mentioned in Section~\ref{sec:data} are general and applicable to any Human-to-Human or Machine-to-Machine dialogues annotated with slot-values. We have yet to evaluate the generalization of our cross-lingual approach across different languages and datasets, and to Human-to-Human dialogues. For instance, we use a Chinese to English translator in this work. Available translation models for low-resource languages have much lower quality, and this will likely lower the performance of this approach.

Another limitation is the lack of human evaluation for agent responses. BLEU score does not correlate well with human judgment, and SER only accounts for the factuality of the response but not the grammaticality or fluency. This problem is also reported in prior works (see Section~\ref{sec:experiments}).
Although finding native speaker evaluators for different languages is a challenge~\cite{pavlick2014language}, in the future, we wish to address this by conducting human evaluations.

\section{Ethical Considerations}
\label{sec:ethics}

We do not foresee any harmful or malicious misuses of the technology developed in this work. The data used to train models is about seeking information about domains like restaurants, hotels and tourist attractions, does not contain any offensive content, and is not unfair or biased against any demographic. This work does focus on two widely-spoken languages, English and Chinese, but we think the cross-lingual approach we proposed can improve future dialogue language technologies for a wider range of languages.

We fine-tune multiple medium-sized (several hundred million parameters) neural networks for our experiments. We took several measures to avoid wasted computation, like performing one run instead of averaging multiple runs (since the numerical difference between different models is large enough), and improving batching and representation that improved training speed, and reduced needed GPU time.
Please refer to Appendix~\ref{sec:implementation} for more details about the amount of computation used in this paper.

\section*{Acknowledgements}
This work is supported in part by the National Science Foundation under Grant No.~1900638, the Alfred P. Sloan Foundation under Grant No.~G-2020-13938, Microsoft, Stanford HAI and the Verdant Foundation.

\bibliography{anthology,custom}

\begin{thebibliography}{42}
\expandafter\ifx\csname natexlab\endcsname\relax\def\natexlab#1{#1}\fi

\bibitem[{Budzianowski et~al.(2018)Budzianowski, Wen, Tseng, Casanueva, Stefan,
  Osman, and Ga{\v{s}}i\'c}]{multiwoz1}
Pawe{\l} Budzianowski, Tsung-Hsien Wen, Bo-Hsiang Tseng, I{\~n}igo Casanueva,
  Ultes Stefan, Ramadan Osman, and Milica Ga{\v{s}}i\'c. 2018.
\newblock Multiwoz - a large-scale multi-domain wizard-of-oz dataset for
  task-oriented dialogue modelling.
\newblock In \emph{Proceedings of the 2018 Conference on Empirical Methods in
  Natural Language Processing (EMNLP)}.

\bibitem[{Campagna et~al.(2019)Campagna, Xu, Moradshahi, Socher, and
  Lam}]{geniepldi19}
Giovanni Campagna, Silei Xu, Mehrad Moradshahi, Richard Socher, and Monica~S.
  Lam. 2019.
\newblock \href {https://doi.org/10.1145/3314221.3314594} {Genie: A generator
  of natural language semantic parsers for virtual assistant commands}.
\newblock In \emph{Proceedings of the 40th ACM SIGPLAN Conference on
  Programming Language Design and Implementation}, PLDI 2019, pages 394--410,
  New York, NY, USA. ACM.

\bibitem[{Chen et~al.(2019)Chen, Chen, Qin, Yan, and
  Wang}]{chen2019semantically}
Wenhu Chen, Jianshu Chen, Pengda Qin, Xifeng Yan, and William~Yang Wang. 2019.
\newblock Semantically conditioned dialog response generation via hierarchical
  disentangled self-attention.
\newblock \emph{arXiv preprint arXiv:1905.12866}.

\bibitem[{Chen et~al.(2018)Chen, Chen, Su, Wang, Yu, Yan, and
  Wang}]{chen-etal-2018-xlnbt}
Wenhu Chen, Jianshu Chen, Yu~Su, Xin Wang, Dong Yu, Xifeng Yan, and
  William~Yang Wang. 2018.
\newblock \href {https://doi.org/10.18653/v1/D18-1038} {{XL}-{NBT}: A
  cross-lingual neural belief tracking framework}.
\newblock In \emph{Proceedings of the 2018 Conference on Empirical Methods in
  Natural Language Processing}, pages 414--424, Brussels, Belgium. Association
  for Computational Linguistics.

\bibitem[{Clark et~al.(2020)Clark, Choi, Collins, Garrette, Kwiatkowski,
  Nikolaev, and Palomaki}]{clark-tydiqa-2020}
Jonathan~H. Clark, Eunsol Choi, Michael Collins, Dan Garrette, Tom Kwiatkowski,
  Vitaly Nikolaev, and Jennimaria Palomaki. 2020.
\newblock \href {https://doi.org/10.1162/tacl_a_00317} {{T}y{D}i {QA}: A
  benchmark for information-seeking question answering in typologically diverse
  languages}.
\newblock \emph{Transactions of the Association for Computational Linguistics},
  8:454--470.

\bibitem[{Devlin et~al.(2018)Devlin, Chang, Lee, and
  Toutanova}]{devlin2018bert}
Jacob Devlin, Ming-Wei Chang, Kenton Lee, and Kristina Toutanova. 2018.
\newblock Bert: Pre-training of deep bidirectional transformers for language
  understanding.
\newblock \emph{arXiv preprint arXiv:1810.04805}.

\bibitem[{Ding et~al.(2021)Ding, Hu, Bing, Aljunied, Joty, Si, and
  Miao}]{globalwoz}
Bosheng Ding, Junjie Hu, Lidong Bing, Sharifah~Mahani Aljunied, Shafiq Joty,
  Luo Si, and Chunyan Miao. 2021.
\newblock \href {http://arxiv.org/abs/2110.07679} {Globalwoz: Globalizing
  multiwoz to develop multilingual task-oriented dialogue systems}.

\bibitem[{Eetemadi and
  Toutanova(2014)}]{eetemadi-toutanova-2014-translationese}
Sauleh Eetemadi and Kristina Toutanova. 2014.
\newblock \href {https://doi.org/10.3115/v1/D14-1018} {Asymmetric features of
  human generated translation}.
\newblock In \emph{Proceedings of the 2014 Conference on Empirical Methods in
  Natural Language Processing ({EMNLP})}, pages 159--164, Doha, Qatar.
  Association for Computational Linguistics.

\bibitem[{Eric et~al.(2019)Eric, Goel, Paul, Sethi, Agarwal, Gao, and
  Hakkani-Tur}]{multiwoz21}
Mihail Eric, Rahul Goel, Shachi Paul, Abhishek Sethi, Sanchit Agarwal, Shuyag
  Gao, and Dilek Hakkani-Tur. 2019.
\newblock Multiwoz 2.1: Multi-domain dialogue state corrections and state
  tracking baselines.
\newblock \emph{arXiv preprint arXiv:1907.01669}.

\bibitem[{faire Linux(2017)}]{num2words}
Savoir faire Linux. 2017.
\newblock num2words.
\newblock \url{https://github.com/savoirfairelinux/num2words}.

\bibitem[{Feng et~al.(2020)Feng, Yang, Cer, Arivazhagan, and
  Wang}]{feng2020language}
Fangxiaoyu Feng, Yinfei Yang, Daniel Cer, Naveen Arivazhagan, and Wei Wang.
  2020.
\newblock Language-agnostic bert sentence embedding.
\newblock \emph{arXiv preprint arXiv:2007.01852}.

\bibitem[{Gao et~al.(2018)Gao, Galley, and Li}]{gao2018neural}
Jianfeng Gao, Michel Galley, and Lihong Li. 2018.
\newblock Neural approaches to conversational ai.
\newblock In \emph{The 41st International ACM SIGIR Conference on Research \&
  Development in Information Retrieval}, pages 1371--1374.

\bibitem[{Gunasekara et~al.(2020)Gunasekara, Kim, D'Haro, Rastogi, Chen, Eric,
  Hedayatnia, Gopalakrishnan, Liu, Huang, Hakkani-Tür, Li, Zhu, Luo, Liden,
  Huang, Shayandeh, Liang, Peng, Zhang, Shukla, Huang, Gao, Mehri, Feng,
  Gordon, Alavi, Traum, Eskenazi, Beirami, Eunjoon, Cho, Crook, De, Geramifard,
  Kottur, Moon, Poddar, and Subba}]{dstc9}
Chulaka Gunasekara, Seokhwan Kim, Luis~Fernando D'Haro, Abhinav Rastogi,
  Yun-Nung Chen, Mihail Eric, Behnam Hedayatnia, Karthik Gopalakrishnan, Yang
  Liu, Chao-Wei Huang, Dilek Hakkani-Tür, Jinchao Li, Qi~Zhu, Lingxiao Luo,
  Lars Liden, Kaili Huang, Shahin Shayandeh, Runze Liang, Baolin Peng, Zheng
  Zhang, Swadheen Shukla, Minlie Huang, Jianfeng Gao, Shikib Mehri, Yulan Feng,
  Carla Gordon, Seyed~Hossein Alavi, David Traum, Maxine Eskenazi, Ahmad
  Beirami, Eunjoon, Cho, Paul~A. Crook, Ankita De, Alborz Geramifard, Satwik
  Kottur, Seungwhan Moon, Shivani Poddar, and Rajen Subba. 2020.
\newblock \href {http://arxiv.org/abs/2011.06486} {Overview of the ninth dialog
  system technology challenge: Dstc9}.

\bibitem[{Hinton et~al.(2015)Hinton, Vinyals, and Dean}]{hinton2015distilling}
Geoffrey Hinton, Oriol Vinyals, and Jeff Dean. 2015.
\newblock \href {http://arxiv.org/abs/1503.02531} {Distilling the knowledge in
  a neural network}.

\bibitem[{Hosseini-Asl et~al.(2020)Hosseini-Asl, McCann, Wu, Yavuz, and
  Socher}]{hosseini2020simple}
Ehsan Hosseini-Asl, Bryan McCann, Chien-Sheng Wu, Semih Yavuz, and Richard
  Socher. 2020.
\newblock A simple language model for task-oriented dialogue.
\newblock \emph{arXiv preprint arXiv:2005.00796}.

\bibitem[{Hung et~al.(2022)Hung, Lauscher, Vuli{\'c}, Ponzetto, and
  Glava{\v{s}}}]{hung2022multi2woz}
Chia-Chien Hung, Anne Lauscher, Ivan Vuli{\'c}, Simone~Paolo Ponzetto, and
  Goran Glava{\v{s}}. 2022.
\newblock Multi2woz: A robust multilingual dataset and conversational
  pretraining for task-oriented dialog.
\newblock \emph{arXiv preprint arXiv:2205.10400}.

\bibitem[{Kale and Rastogi(2020)}]{t2g2}
Mihir Kale and Abhinav Rastogi. 2020.
\newblock \href {https://doi.org/10.18653/v1/2020.emnlp-main.527} {Template
  guided text generation for task-oriented dialogue}.
\newblock In \emph{Proceedings of the 2020 Conference on Empirical Methods in
  Natural Language Processing (EMNLP)}, pages 6505--6520, Online. Association
  for Computational Linguistics.

\bibitem[{Kingma and Ba(2014)}]{kingma2014adam}
Diederik~P Kingma and Jimmy Ba. 2014.
\newblock Adam: A method for stochastic optimization.
\newblock \emph{arXiv preprint arXiv:1412.6980}.

\bibitem[{Kudo and Richardson(2018)}]{kudo2018sentencepiece}
Taku Kudo and John Richardson. 2018.
\newblock Sentencepiece: A simple and language independent subword tokenizer
  and detokenizer for neural text processing.
\newblock \emph{arXiv preprint arXiv:1808.06226}.

\bibitem[{Lei et~al.(2018)Lei, Jin, Kan, Ren, He, and Yin}]{lei2018sequicity}
Wenqiang Lei, Xisen Jin, Min-Yen Kan, Zhaochun Ren, Xiangnan He, and Dawei Yin.
  2018.
\newblock Sequicity: Simplifying task-oriented dialogue systems with single
  sequence-to-sequence architectures.
\newblock In \emph{Proceedings of the 56th Annual Meeting of the Association
  for Computational Linguistics (Volume 1: Long Papers)}, pages 1437--1447.

\bibitem[{Li et~al.(2021{\natexlab{a}})Li, Arora, Chen, Gupta, Gupta, and
  Mehdad}]{mtop}
Haoran Li, Abhinav Arora, Shuohui Chen, Anchit Gupta, Sonal Gupta, and Yashar
  Mehdad. 2021{\natexlab{a}}.
\newblock \href {https://doi.org/10.18653/v1/2021.eacl-main.257} {{MTOP}: A
  comprehensive multilingual task-oriented semantic parsing benchmark}.
\newblock In \emph{Proceedings of the 16th Conference of the European Chapter
  of the Association for Computational Linguistics: Main Volume}, pages
  2950--2962, Online. Association for Computational Linguistics.

\bibitem[{Li et~al.(2021{\natexlab{b}})Li, Zhu, Luo, Liden, Huang, Shayandeh,
  Liang, Peng, Zhang, Shukla, Takanobu, Huang, and Gao}]{li2021multi-domain}
Jinchao Li, Qi~Zhu, Lingxiao Luo, Lars Liden, Kaili Huang, Shahin Shayandeh,
  Runze Liang, Baolin Peng, Zheng Zhang, Swadheen Shukla, Ryuichi Takanobu,
  Minlie Huang, and Jianfeng Gao. 2021{\natexlab{b}}.
\newblock \href
  {https://www.microsoft.com/en-us/research/publication/multi-domain-task-oriented-dialog-challenge-ii-at-dstc9/}
  {Multi-domain task-oriented dialog challenge ii at dstc9}.
\newblock In \emph{AAAI-2021 Dialog System Technology Challenge 9 Workshop}.

\bibitem[{Lin et~al.(2020)Lin, Madotto, Winata, and Fung}]{lin-etal-2020-mintl}
Zhaojiang Lin, Andrea Madotto, Genta~Indra Winata, and Pascale Fung. 2020.
\newblock \href {https://doi.org/10.18653/v1/2020.emnlp-main.273} {{M}in{TL}:
  Minimalist transfer learning for task-oriented dialogue systems}.
\newblock In \emph{Proceedings of the 2020 Conference on Empirical Methods in
  Natural Language Processing (EMNLP)}, pages 3391--3405, Online. Association
  for Computational Linguistics.

\bibitem[{Lin et~al.(2021)Lin, Madotto, Winata, Xu, Jiang, Hu, Shi, and
  Fung}]{lin2021bitod}
Zhaojiang Lin, Andrea Madotto, Genta~Indra Winata, Peng Xu, Feijun Jiang,
  Yuxiang Hu, Chen Shi, and Pascale Fung. 2021.
\newblock {B}i{T}o{D}: A bilingual multi-domain dataset for task-oriented
  dialogue modeling.
\newblock \emph{Proceedings of the Neural Information Processing Systems Track
  on Datasets and Benchmarks 1 pre-proceedings (NeurIPS Datasets and Benchmarks
  2021)}.

\bibitem[{Liu et~al.(2020)Liu, Gu, Goyal, Li, Edunov, Ghazvininejad, Lewis, and
  Zettlemoyer}]{liu2020multilingual}
Yinhan Liu, Jiatao Gu, Naman Goyal, Xian Li, Sergey Edunov, Marjan
  Ghazvininejad, Mike Lewis, and Luke Zettlemoyer. 2020.
\newblock \href {http://arxiv.org/abs/2001.08210} {Multilingual denoising
  pre-training for neural machine translation}.

\bibitem[{Moradshahi et~al.(2020)Moradshahi, Campagna, Semnani, Xu, and
  Lam}]{moradshahi-etal-2020-localizing}
Mehrad Moradshahi, Giovanni Campagna, Sina Semnani, Silei Xu, and Monica Lam.
  2020.
\newblock \href {https://doi.org/10.18653/v1/2020.emnlp-main.481} {Localizing
  open-ontology {QA} semantic parsers in a day using machine translation}.
\newblock In \emph{Proceedings of the 2020 Conference on Empirical Methods in
  Natural Language Processing (EMNLP)}, pages 5970--5983, Online. Association
  for Computational Linguistics.

\bibitem[{Moradshahi et~al.(2023)Moradshahi, Tsai, Campagna, and
  Lam}]{moradshahi2021contextual}
Mehrad Moradshahi, Victoria Tsai, Giovanni Campagna, and Monica~S Lam. 2023.
\newblock Contextual semantic parsing for multilingual task-oriented dialogues.
\newblock In \emph{Proceedings of the European Chapter of the Association for
  Computational Linguistics (EACL)}.

\bibitem[{Mrk{\v{s}}i{\'c} et~al.(2017)Mrk{\v{s}}i{\'c}, Vuli{\'c},
  {\'O}~S{\'e}aghdha, Leviant, Reichart, Ga{\v{s}}i{\'c}, Korhonen, and
  Young}]{mrksic-etal-2017-semantic}
Nikola Mrk{\v{s}}i{\'c}, Ivan Vuli{\'c}, Diarmuid {\'O}~S{\'e}aghdha, Ira
  Leviant, Roi Reichart, Milica Ga{\v{s}}i{\'c}, Anna Korhonen, and Steve
  Young. 2017.
\newblock \href {https://doi.org/10.1162/tacl_a_00063} {Semantic specialization
  of distributional word vector spaces using monolingual and cross-lingual
  constraints}.
\newblock \emph{Transactions of the Association for Computational Linguistics},
  5:309--324.

\bibitem[{Papineni et~al.(2002)Papineni, Roukos, Ward, and
  Zhu}]{papineni2002bleu}
Kishore Papineni, Salim Roukos, Todd Ward, and Wei-Jing Zhu. 2002.
\newblock Bleu: a method for automatic evaluation of machine translation.
\newblock In \emph{Proceedings of the 40th annual meeting on association for
  computational linguistics}, pages 311--318. Association for Computational
  Linguistics.

\bibitem[{Paszke et~al.(2019)Paszke, Gross, Massa, Lerer, Bradbury, Chanan,
  Killeen, Lin, Gimelshein, Antiga et~al.}]{paszke2019pytorch}
Adam Paszke, Sam Gross, Francisco Massa, Adam Lerer, James Bradbury, Gregory
  Chanan, Trevor Killeen, Zeming Lin, Natalia Gimelshein, Luca Antiga, et~al.
  2019.
\newblock Pytorch: An imperative style, high-performance deep learning library.
\newblock \emph{Advances in neural information processing systems},
  32:8026--8037.

\bibitem[{Pavlick et~al.(2014)Pavlick, Post, Irvine, Kachaev, and
  Callison-Burch}]{pavlick2014language}
Ellie Pavlick, Matt Post, Ann Irvine, Dmitry Kachaev, and Chris Callison-Burch.
  2014.
\newblock The language demographics of amazon mechanical turk.
\newblock \emph{Transactions of the Association for Computational Linguistics},
  2:79--92.

\bibitem[{Quan et~al.(2020)Quan, Zhang, Cao, Li, and Xiong}]{risawoz}
Jun Quan, Shian Zhang, Qian Cao, Zizhong Li, and Deyi Xiong. 2020.
\newblock \href {https://doi.org/10.18653/v1/2020.emnlp-main.67} {{R}i{SAWOZ}:
  A large-scale multi-domain {W}izard-of-{O}z dataset with rich semantic
  annotations for task-oriented dialogue modeling}.
\newblock In \emph{Proceedings of the 2020 Conference on Empirical Methods in
  Natural Language Processing (EMNLP)}, pages 930--940, Online. Association for
  Computational Linguistics.

\bibitem[{Ramadan et~al.(2018)Ramadan, Budzianowski, and Gasic}]{multiwoz2}
Osman Ramadan, Pawe{\l} Budzianowski, and Milica Gasic. 2018.
\newblock Large-scale multi-domain belief tracking with knowledge sharing.
\newblock In \emph{Proceedings of the 56th Annual Meeting of the Association
  for Computational Linguistics}, volume~2, pages 432--437.

\bibitem[{Razumovskaia et~al.(2021)Razumovskaia, Glava{\v{s}}, Majewska,
  Korhonen, and Vulic}]{razumovskaia2021crossing}
Evgeniia Razumovskaia, Goran Glava{\v{s}}, Olga Majewska, Anna Korhonen, and
  Ivan Vulic. 2021.
\newblock Crossing the conversational chasm: A primer on multilingual
  task-oriented dialogue systems.
\newblock \emph{arXiv preprint arXiv:2104.08570}.

\bibitem[{Schuster et~al.(2019)Schuster, Gupta, Shah, and
  Lewis}]{schuster-etal-2019-cross-lingual}
Sebastian Schuster, Sonal Gupta, Rushin Shah, and Mike Lewis. 2019.
\newblock \href {https://doi.org/10.18653/v1/N19-1380} {Cross-lingual transfer
  learning for multilingual task oriented dialog}.
\newblock In \emph{Proceedings of the 2019 Conference of the North {A}merican
  Chapter of the Association for Computational Linguistics: Human Language
  Technologies, Volume 1 (Long and Short Papers)}, pages 3795--3805,
  Minneapolis, Minnesota. Association for Computational Linguistics.

\bibitem[{Scrapinghub(2015)}]{dateparser}
Scrapinghub. 2015.
\newblock dateparser.
\newblock \url{https://github.com/scrapinghub/dateparser}.

\bibitem[{Sherborne et~al.(2020)Sherborne, Xu, and
  Lapata}]{sherborne-etal-2020-bootstrapping}
Tom Sherborne, Yumo Xu, and Mirella Lapata. 2020.
\newblock \href {https://doi.org/10.18653/v1/2020.findings-emnlp.45}
  {Bootstrapping a crosslingual semantic parser}.
\newblock In \emph{Findings of the Association for Computational Linguistics:
  EMNLP 2020}, pages 499--517, Online. Association for Computational
  Linguistics.

\bibitem[{Wen et~al.(2015)Wen, Gasic, Mrksic, Su, Vandyke, and
  Young}]{wen2015semantically}
Tsung-Hsien Wen, Milica Gasic, Nikola Mrksic, Pei-Hao Su, David Vandyke, and
  Steve Young. 2015.
\newblock Semantically conditioned lstm-based natural language generation for
  spoken dialogue systems.
\newblock \emph{arXiv preprint arXiv:1508.01745}.

\bibitem[{Wolf et~al.(2019)Wolf, Debut, Sanh, Chaumond, Delangue, Moi, Cistac,
  Rault, Louf, Funtowicz, and Brew}]{Wolf2019HuggingFacesTS}
Thomas Wolf, Lysandre Debut, Victor Sanh, Julien Chaumond, Clement Delangue,
  Anthony Moi, Pierric Cistac, Tim Rault, R'emi Louf, Morgan Funtowicz, and
  Jamie Brew. 2019.
\newblock Huggingface's transformers: State-of-the-art natural language
  processing.
\newblock \emph{ArXiv}, abs/1910.03771.

\bibitem[{Xue et~al.(2021)Xue, Constant, Roberts, Kale, Al-Rfou, Siddhant,
  Barua, and Raffel}]{mt5}
Linting Xue, Noah Constant, Adam Roberts, Mihir Kale, Rami Al-Rfou, Aditya
  Siddhant, Aditya Barua, and Colin Raffel. 2021.
\newblock \href {https://doi.org/10.18653/v1/2021.naacl-main.41} {mt5: A
  massively multilingual pre-trained text-to-text transformer}.
\newblock \emph{Proceedings of the 2021 Conference of the North American
  Chapter of the Association for Computational Linguistics: Human Language
  Technologies}.

\bibitem[{Zhu et~al.(2020)Zhu, Huang, Zhang, Zhu, and
  Huang}]{zhu-etal-2020-crosswoz}
Qi~Zhu, Kaili Huang, Zheng Zhang, Xiaoyan Zhu, and Minlie Huang. 2020.
\newblock \href {https://doi.org/10.1162/tacl_a_00314} {{C}ross{WOZ}: A
  large-scale {C}hinese cross-domain task-oriented dialogue dataset}.
\newblock \emph{Transactions of the Association for Computational Linguistics},
  8:281--295.

\bibitem[{Zuo et~al.(2021)Zuo, Qian, Yang, and Yu}]{zuo2021allwoz}
Lei Zuo, Kun Qian, Bowen Yang, and Zhou Yu. 2021.
\newblock Allwoz: Towards multilingual task-oriented dialog systems for all.
\newblock \emph{arXiv preprint arXiv:2112.08333}.

\end{thebibliography}
\bibliographystyle{acl_natbib}

\newpage
\appendix
\section{Appendix}
\label{sec:appendix}

\subsection{Dialogue Examples}
We include the same example from BiToD's English validation set both in our \repr representation (Table~\ref{tab:examples-ours})
and in the original (Table~\ref{tab:examples-original}) representation, along with model predictions in the full-shot setting. For brevity, only the first 3 turns are shown.

In Table~\ref{tab:examples-original}, we observe that the model fails to ask for the hotel price-range in the second turn and makes an API call instead. Since the API call results are carried over between turns in this representation, in the third turn, the model sees those results in the input and falsely assumes it does not need to make an API call anymore, ultimately resulting in an incorrect response. Compare this to our representation in Table~\ref{tab:examples-ours}. This example shows the importance of separation between API call detection and response generation.

Another phenomenon we often observe is that the model asks for more information than it should according to the gold agent dialogue act. As shown in Table~\ref{tab:examples-ours}, in the second turn, the agent requests user to provide the desired location for the hotel as well as the price range. We believe the main reason for this behavior is the randomness in the agent policy of the BiToD's dialogue simulator. For example, if the agent needs to fill out two slots to make an API call, it can do so by requesting both in the same turn, or one turn at a time. This behavior, though reasonable, is penalized during evaluation, and predictions are considered incorrect if they contain extraneous slots.


\begin{table*}
\begin{tabularx}{\linewidth}{c|c|c|X}
\specialrule{.1em}{.05em}{.05em} 
\multirow{17}{*}{\bf Turn 1} & \multirow{4}{*}{DST} & \multirow{2}{*}{Input} &  \small DST: <state> null <endofstate> <history> USER: I'd like hotel recommendations. <endofhistory> \\
\cline{3-4}
& & Target & \small ( hotels search ) \\
\cline{3-4}
& & Prediction & \small ( hotels search ) \\
\cline{2-4}
 & \multirow{4}{*}{ACD} & \multirow{2}{*}{Input} &  \small API: <knowledge> null <endofknowledge> <state> ( hotels search ) <endofstate> <history> USER: I'd like hotel recommendations. <endofhistory> \\
\cline{3-4}
& & Target & \small no \\
\cline{3-4}
& & Prediction & \small no \\
\cline{2-4}
 & \multirow{4}{*}{DAG} & \multirow{2}{*}{Input} &  \small ACTS: <knowledge> null <endofknowledge> <state> ( hotels search ) <endofstate> <history> USER: I'd like hotel recommendations. <endofhistory> \\
\cline{3-4}
& & Target & \small ( hotels search ) request rating , request stars \\
\cline{3-4}
& & Prediction & \small ( hotels search ) request rating , request stars \\
\cline{2-4}
 & \multirow{5}{*}{RG} & \multirow{2}{*}{Input} &  \small RG: <actions> ( hotels search ) request rating , request stars <endofactions> <history> USER: I'd like hotel recommendations. <endofhistory> \\
\cline{3-4}
& & Target & \small Certainly. Do you have any requirements for the hotel's rating or the number of stars of the hotel? \\
\cline{3-4}
& & Prediction & \small Do you have a preference on how many stars and what rating the hotel should have? \\
\specialrule{.1em}{.05em}{.05em} 
\multirow{16}{*}{\bf Turn 2} & \multirow{4}{*}{DST} & \multirow{2}{*}{Input} &  \small DST: <state> ( hotels search ) <endofstate> <history> AGENT\_ACTS: ( hotels search ) request rating , request stars USER: The rating doesn't matter, but should be at least 5 stars. <endofhistory> \\
\cline{3-4}
& & Target & \small ( hotels search ) rating equal\_to " don't care " , stars at\_least " 5 " \\
\cline{3-4}
& & Prediction & \small ( hotels search ) rating equal\_to " don't care " , stars at\_least " 5 " \\
\cline{2-4}
 & \multirow{4}{*}{ACD} & \multirow{2}{*}{Input} &  \small API: <knowledge> null <endofknowledge> <state> ( hotels search ) rating equal\_to " don't care " , stars at\_least " 5 " <endofstate> <history> AGENT\_ACTS: ( hotels search ) request rating , request stars USER: The rating doesn't matter, but should be at least 5 stars. <endofhistory> \\
\cline{3-4}
& & Target & \small no \\
\cline{3-4}
& & Prediction & \small no \\
\cline{2-4}
 & \multirow{5}{*}{DAG} & \multirow{2}{*}{Input} &  \small ACTS: <knowledge> null <endofknowledge> <state> ( hotels search ) rating equal\_to " don't care " , stars at\_least " 5 " <endofstate> <history> AGENT\_ACTS: ( hotels search ) request rating , request stars USER: The rating doesn't matter, but should be at least 5 stars. <endofhistory> \\
\cline{3-4}
& & Target & \small ( hotels search ) request price\_level \\
\cline{3-4}
& & Prediction & \small ( hotels search ) request location , request price\_level \\
\cline{2-4}
 & \multirow{4}{*}{RG} & \multirow{2}{*}{Input} &  \small RG: <actions> ( hotels search ) request price\_level <endofactions> <history> USER: The rating doesn't matter, but should be at least 5 stars. <endofhistory> \\
\cline{3-4}
& & Target & \small Do you have a price range for the hotel? \\
\cline{3-4}
& & Prediction & \small And what about location? Do you have a price range for the hotel? \\
\specialrule{.1em}{.05em}{.05em} 
\end{tabularx}
\end{table*}

\begin{table*}
\begin{tabularx}{\linewidth}{c|c|c|X}
\specialrule{.1em}{.05em}{.05em} 
\multirow{21}{*}{\bf Turn 3} & \multirow{4}{*}{DST} & \multirow{2}{*}{Input} &  \small DST: <state> ( hotels search ) rating equal\_to " don't care " , stars at\_least " 5 " <endofstate> <history> AGENT\_ACTS\_PREV: ( hotels search ) request rating , request stars AGENT\_ACTS: ( hotels search ) request price\_level USER: cheap <endofhistory> \\
\cline{3-4}
& & Target & \small ( hotels search ) price\_level equal\_to " cheap " \\
\cline{3-4}
& & Prediction & \small ( hotels search ) price\_level equal\_to " cheap " \\
\cline{2-4}
 & \multirow{5}{*}{ACD} & \multirow{3}{*}{Input} &  \small API: <knowledge> null <endofknowledge> <state> ( hotels search ) price\_level equal\_to " cheap " , rating equal\_to " don't care " , stars at\_least " 5 " <endofstate> <history> AGENT\_ACTS\_PREV: ( hotels search ) request rating , request stars AGENT\_ACTS: ( hotels search ) request price\_level USER: cheap <endofhistory> \\
\cline{3-4}
& & Target & \small yes \\
\cline{3-4}
& & Prediction & \small yes \\
\cline{2-4}
 & \multirow{8}{*}{DAG} & \multirow{5}{*}{Input} &  \small ACTS: <knowledge> ( hotels search ) available\_options " 4 " , location " Mong Kok | Kowloon | Yau Tsim Mong District " , name " Royal Plaza Hotel " , price\_level " cheap " , price\_per\_night " 793 HKD " , rating " 9 " , stars " 5 " <endofknowledge> <state> ( hotels search ) price\_level equal\_to " cheap " , rating equal\_to " don't care " , stars at\_least " 5 " <endofstate> <history> AGENT\_ACTS\_PREV: ( hotels search ) request rating , request stars AGENT\_ACTS: ( hotels search ) request price\_level USER: cheap <endofhistory> \\
\cline{3-4}
& & Target & \small ( hotels search ) offer available\_options equal\_to " 4 " , offer name equal\_to " Royal Plaza Hotel " , offer rating equal\_to " 9 " \\
\cline{3-4}
& & Prediction & \small ( hotels search ) offer available\_options equal\_to " 4 " , offer name equal\_to " Royal Plaza Hotel " , offer rating equal\_to " 9 " \\
\cline{2-4}
 & \multirow{5}{*}{RG} & \multirow{2}{*}{Input} &  \small RG: <actions> ( hotels search ) offer available\_options equal\_to " 4 " , offer name equal\_to " Royal Plaza Hotel " , offer rating equal\_to " 9 " <endofactions> <history> USER: cheap <endofhistory> \\
\cline{3-4}
& & Target & \small Okay. There are 4 hotels available. I recommend the Royal Plaza Hotel, which has a 9 rating. \\
\cline{3-4}
& & Prediction & \small There are 4 available hotels. I recommend Royal Plaza Hotel. Its rating is 9. \\
\specialrule{.1em}{.05em}{.05em} 
\end{tabularx}
\caption{An example from BiToD English validation set in \repr representation, along with our mBART model predictions. For brevity, only the first 3 turns are shown.}
\label{tab:examples-ours}
\end{table*}

\begin{table*}
\begin{tabularx}{\linewidth}{c|c|c|X}
\specialrule{.1em}{.05em}{.05em} 
\multirow{8}{*}{\bf Turn 1} & \multirow{4}{*}{DST} & \multirow{2}{*}{Input} &  \small Track Dialogue State:<knowledge><dialogue\_state> <user> I'd like hotel recommendations. \\
\cline{3-4}
& & Target & \small <API> hotels search \\
\cline{3-4}
& & Prediction & \small <API> hotels search \\
\cline{2-4}
 & \multirow{4}{*}{API/ \textbf{Response}} & \multirow{2}{*}{Input} &  \small Generate Response:<knowledge><dialogue\_state> <API> hotels search<user> I'd like hotel recommendations. \\
\cline{3-4}
& & Target & \small Certainly. Do you have any requirements for the hotel's rating or the number of stars of the hotel? \\
\cline{3-4}
& & Prediction & \small What rating would you like the hotel to have? \\
\specialrule{.1em}{.05em}{.05em} 
\multirow{13}{*}{\bf Turn 2} & \multirow{6}{*}{DST} & \multirow{2}{*}{Input} &  \small Track Dialogue State:<knowledge><dialogue\_state> <API> hotels search<user> I'd like hotel recommendations.<system> Certainly. Do you have any requirements for the hotel's rating or the number of stars of the hotel?<user> The rating doesn't matter, but should be at least 5 stars. \\
\cline{3-4}
& & Target & \small <API> hotels search<slot> rating<relation> equal\_to<value> don't care<slot> stars<relation> at\_least<value> 5 \\
\cline{3-4}
& & Prediction & \small <API> hotels search<slot> rating<relation> equal\_to<value> don't care<slot> stars<relation> at\_least<value> 5 \\
\cline{2-4}
 & \multirow{7}{*}{API/ \textbf{Response}} & \multirow{2}{*}{Input} &  \small Generate Response:<knowledge><dialogue\_state> <API> hotels search<slot> rating<relation> equal\_to<value> don't care<slot> stars<relation> at\_least<value> 5<user> I'd like hotel recommendations.<system> Certainly. Do you have any requirements for the hotel's rating or the number of stars of the hotel?<user> The rating doesn't matter, but should be at least 5 stars. \\
\cline{3-4}
& & Target & \small Do you have a price range for the hotel? \\
\cline{3-4}
& & Prediction & \small I found 43 hotels. I recommend Cordis, Hong Kong, which has a rating of 10. \\
\specialrule{.1em}{.05em}{.05em} 
\multirow{18}{*}{\bf Turn 3} & \multirow{6}{*}{DST} & \multirow{4}{*}{Input} &  \small Track Dialogue State:<knowledge><dialogue\_state> <API> hotels search<slot> rating<relation> equal\_to<value> don't care<slot> stars<relation> at\_least<value> 5<user> The rating doesn't matter, but should be at least 5 stars.<system> Do you have a price range for the hotel?<user> cheap \\
\cline{3-4}
& & Target & \small <API> hotels search<slot> price\_level<relation> equal\_to<value> cheap \\
\cline{3-4}
& & Prediction & \small <API> hotels search<slot> price\_level<relation> equal\_to<value> cheap \\
\cline{2-4}
 & \multirow{6}{*}{\textbf{API}/ Response} & \multirow{4}{*}{Input} &  \small Generate Response:<knowledge><dialogue\_state> <API> hotels search<slot> rating<relation> equal\_to<value> don't care<slot> stars<relation> at\_least<value> 5<slot> price\_level<relation> equal\_to<value> cheap<user> The rating doesn't matter, but should be at least 5 stars.<system> Do you have a price range for the hotel?<user> cheap \\
\cline{3-4}
& & Target & \small <API> hotels search \\
\cline{3-4}
& & Prediction & \small -- \\
\cline{2-4}
 & \multirow{10}{*}{API/ \textbf{Response}} & \multirow{7}{*}{Input} &  \small Generate Response:<knowledge> [hotels]<slot> name<value> Royal Plaza Hotel<slot> location<value> Mong Kok<value> Kowloon<value> Yau Tsim Mong District<slot> price\_level<value> cheap<slot> price\_per\_night<value> 793 HKD<slot> rating<value> 9<slot> stars<value> 5<slot> available\_options<value> 4<dialogue\_state> <API> hotels search<slot> rating<relation> equal\_to<value> don't care<slot> stars<relation> at\_least<value> 5<slot> price\_level<relation> equal\_to<value> cheap<user> The rating doesn't matter, but should be at least 5 stars.<system> Do you have a price range for the hotel?<user> cheap<API> hotels search \\
\cline{3-4}
& & Target & \small Okay. There are 4 hotels available. I recommend the Royal Plaza Hotel, which has a 9 rating. \\
\cline{3-4}
& & Prediction & \small  The hotel costs 839 HKD per night. \\
\specialrule{.1em}{.05em}{.05em} 
\end{tabularx}
\caption{Same example as in Table~\ref{tab:examples-ours} but in the original representation from \citet{lin2021bitod}, along with MinTL(mT5) model predictions. For brevity, only the first 3 turns are shown.}
\label{tab:examples-original}
\end{table*}

 

\end{document}